\documentclass[conference]{IEEEtran}
\usepackage{amsmath}
\usepackage{cases}
\usepackage{mathtools}
\usepackage{times}
\usepackage{graphicx}
\usepackage{graphics}
\usepackage{mathptmx}
\usepackage{times}
\usepackage{amssymb}
\usepackage{dcolumn}
\usepackage{bm}
\usepackage{units}

\begin{document}
\title{Fatiguing STDP: Learning from Spike-Timing Codes in the Presence of Rate Codes}
\author{
\IEEEauthorblockN{Timoleon Moraitis, Abu Sebastian, Irem Boybat,  Manuel Le Gallo, Tomas Tuma, and Evangelos Eleftheriou} \IEEEauthorblockA{IBM Research -- Zurich, 8803 R\"{u}schlikon, Switzerland \\ Email: ora@zurich.ibm.com, ase@zurich.ibm.com} } \maketitle

\begin{abstract}
Spiking neural networks (SNNs) could play a key role in unsupervised machine learning applications, by virtue of strengths related to learning from the fine temporal structure of event-based signals. However, some spike-timing-related strengths of SNNs are hindered by the sensitivity of spike-timing-dependent plasticity (STDP) rules to input spike rates, as fine temporal correlations may be obstructed by coarser correlations between firing rates. In this article, we propose a spike-timing-dependent learning rule that allows a neuron to learn from the temporally-coded information despite the presence of rate codes. Our long-term plasticity rule makes use of short-term synaptic fatigue dynamics. We show analytically that, in contrast to conventional STDP rules, our fatiguing STDP (FSTDP) helps learn the temporal code, and we derive the necessary conditions to optimize the learning process. We showcase the effectiveness of FSTDP in learning spike-timing correlations among processes of different rates in synthetic data. Finally, we use FSTDP to detect correlations in real-world weather data from the United States in an experimental realization of the algorithm that uses a neuromorphic hardware platform comprising phase-change memristive devices. Taken together, our analyses and demonstrations suggest that FSTDP paves the way for the exploitation of the spike-based strengths of SNNs in real-world applications.
\end{abstract}

\IEEEpeerreviewmaketitle

\section{Introduction}
Spiking neural networks (SNNs) are algorithms that underlie the brain's function, and their artificial implementations are viewed as an emerging generation of artificial neural networks (ANNs) that could enable efficient solutions for tasks of machine intelligence. As evidence of SNNs' potential, previous generations of ANNs that are more loosely based on biological ones have recently exploited the increases in available computational resources and large annotated datasets to provide record-breaking performance in several machine-learning tasks such as image and speech recognition, natural-language processing, playing complex games, and data analytics for scientific or business purposes \cite{Y2015lecunNature}. These conventional ANNs achieve their most impressive results by following supervised learning rules and by receiving their input and communicating between neurons in discrete time-steps and in real-valued signals.

In contrast to these networks, SNNs in the brain learn in a largely unsupervised way, as they need to uncover the structure of natural data provided without labels. Despite the lack of supervision, biological SNNs perform a variety of tasks at levels yet unsurpassed by unsupervised ANN systems. A key characteristic that underlies this difference in performance is that the input and the communication in SNNs are handled by streams of binary events called spikes. This allows, first, time to represent itself in the relative timing of events, second, the massively distributed neuronal and synaptic computations to be performed asynchronously, and third, the network to exploit the full information that is embedded in the fine structure of the spike timing sequences, apart from their mean rate. In addition, artificial SNNs are implementable in compact, low-power neuromorphic hardware. Hence, SNNs and their physical implementations are being actively researched as an alternative to conventional ANNs \cite{Y1997maassNN, Y2014merollaScience, Y2015indiveriIEEEProc}.

\begin{figure}[h!]
\centering
\begin{tabular}{c}
\includegraphics[width = 3.5in]{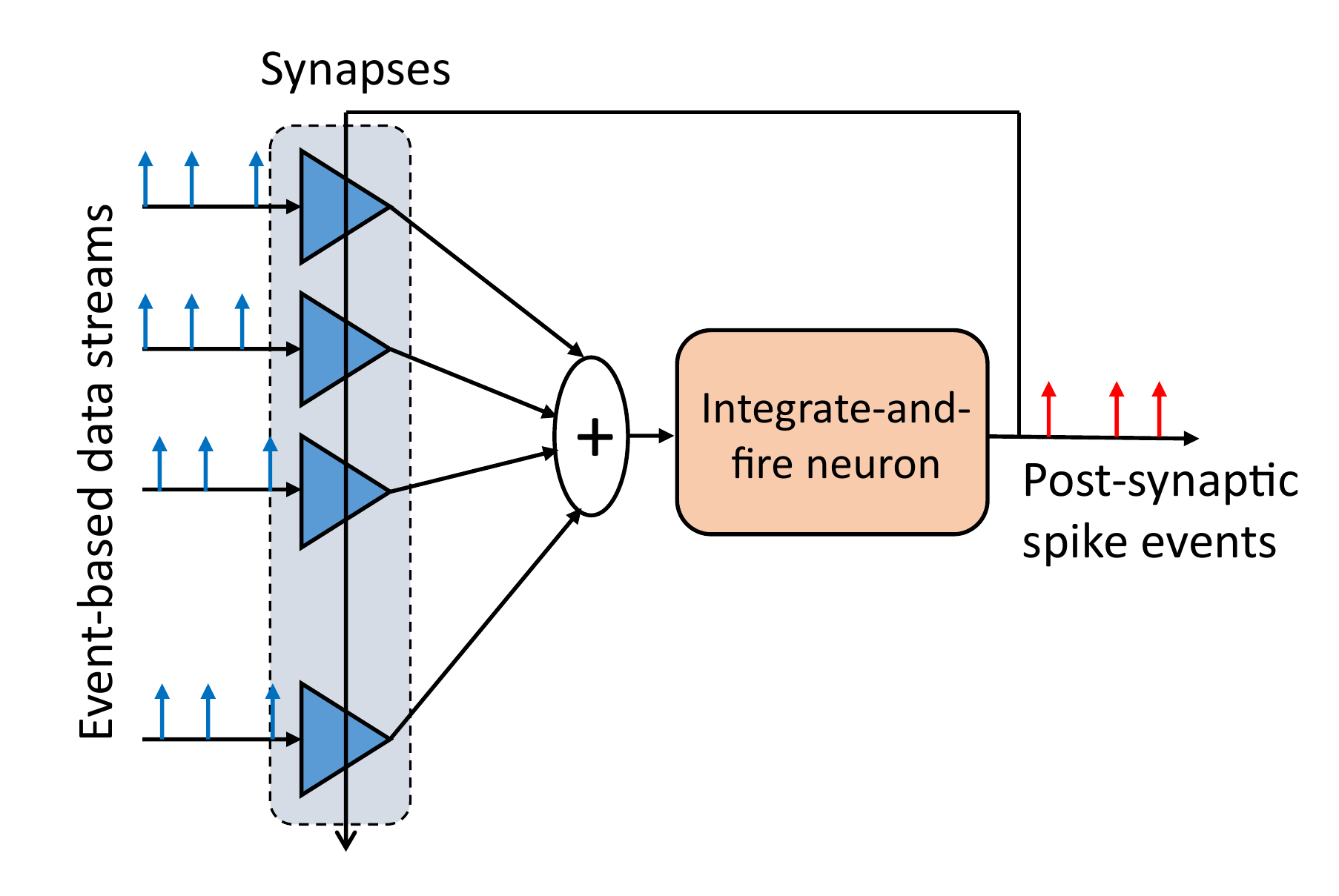}
\end{tabular}
\caption{An integrate-and-fire neuron equipped with an array of plastic synapses and an STDP learning rule can be used to detect temporal
correlations between event-based data streams. The correlations between the individual input streams can be inferred from the steady-state
distribution of the synaptic weights or the firing activity of the neuron.} \label{fig:corr}
\end{figure}

SNNs could play a central role in the domain of unsupervised learning, which is a key computational task in data analytics, either as a stand-alone technique or as a pre-training phase for subsequent supervised learning, because many of today's big-data sources lack the labeled samples and reliable training sets needed for supervised learning. The information carried by the temporal structure in streams of events that occur naturally or are generated by event-based biological or artificial monitoring systems, can be exploited in SNNs by unsupervised spike-timing-dependent plasticity (STDP) rules that are based on the temporal relationship between the postsynaptic activity, i.e., postsynaptic spikes and related variables, and the presynaptic input. For instance, an integrate-and-fire neuron can be used to learn from temporal correlations between event-based data streams \cite{Y2016tumaNatNano, Y2003gutigJNeuro}. As shown in Fig. \ref{fig:corr}, each event arrives as a spike at a corresponding synapse, and a postsynaptic potential is generated and added to the membrane potential of the neuron. The temporal correlations between the presynaptic input spikes and the neuronal firing events result in an evolution of the synaptic weights due to a feedback-driven competition among the synapses. In the steady state, the correlations between the individual input streams can be inferred from the distribution of the synaptic weights or the resulting firing activity of the postsynaptic neuron.

One key problem in fully exploiting the spike-timing-related strengths of SNNs is that STDP algorithms are highly sensitive to the rate of the input spikes in addition to the relative timing of individual spikes between the data streams \cite{Kempter1999, Gilson2011}. If simultaneously to the temporal coding the input streams carry additional information that is rate-coded, i.e., carried by the rate of occurrence of events, the learning of the temporally-coded information may be obstructed. Thus, detecting temporal correlations in the presence of rate codes is a challenge to be addressed.

To address this problem, we present a learning rule that combines the long-term STDP dynamics with a mechanism of short-term synaptic
"fatigue" dynamics. First, we motivate and present a description of the proposed learning rule which we call \textit{fatiguing STDP} or \textit{FSTDP}.
Subsequently, we present an analytical study of how FSTDP enables learning of spike-timing codes when STDP does not. We present a simulation study using
synthetic data that shows the effectiveness of FSTDP in detecting low-rate correlated Poisson processes among processes of different rates. Finally, we illustrate the concept on real-world event-based rainfall data from weather stations of the United States of America. We show how FSTDP can be used to learn temporal
correlations in spite of the significant differences in the rate of rainfall across stations. We demonstrate an experimental realization of such an
algorithm with a memristive neuromorphic platform in which the synaptic weights are stored and modified \textit{in-situ} in phase-change memory devices.

\section{Fatiguing spike-timing-dependent plasticity}
\label{sec:fstdp}
In this section, we start by formalizing the task of learning spike-timing codes and the problem of coexisting rate codes to motivate the learning rule we propose, and then introduce and describe the rule.

Learning a task from a spike-timing code consists in learning to detect the temporal correlations across the streams that carry the code. Therefore, in an SNN, the neurons need to learn to associate the strongly covarying inputs. This can be well illustrated as the discovery of clusters of high covariance $c_{ij}$ in the covariance matrix of the input synapse pairs $i, j$. In STDP, a neuron approximates this by potentiating, i.e., strengthening, and depressing, i.e., weakening, the synapses $i$ whose input shows high and low covariance $c_{iN}$ respectively with the neuron's $N$ own output, i.e. the weighted sum of the input streams. This is a good approximation because inputs that covary strongly with the neuron's output which they drive likely covary strongly with one another. So covariances $c_{ij}$ between inputs define the learning task, and covariances $c_{iN}$ of the inputs with the sum of all inputs set a prediction for the neuron's optimal approximation to the task.

The neuron's STDP synapses compare the inputs directly with the neuron's output, so a neuron's learning is sensitive specifically to the uncentered covariance of the inputs, given by $cov(X_i,X_j)=E[X_i\cdot X_j]$ for two input streams $X_i$ and $X_j$. If in addition to the covariances introduced by the correlations in timings of individual spikes there are covariances introduced by correlations in the rates of the inputs, these rate-induced covariances dominate the uncentered covariance, because of the spurious correlations of individual spike timings that are added by the slower covarying rates. To detect the fast covariances in the presence of the slow ones, individual spikes from high-rate channels must contribute less to the computed covariance than those from low-rate channels. Based on this, we introduce the measure of covariance normalized by the means of the inputs, which in the case of spike trains are the mean firing rates. We call this measure normalized covariance:
\begin{equation}
	normcov(X_i,X_j)=E\left[\frac{X_i}{E[X_i]}\cdot\frac{X_j}{E[X_j]}\right].
\end{equation}

Therefore, for a neuron to learn a spike-timing code in the presence of a rate code, a modified STDP rule is needed. This rule should approximate the learning of covariances of rate-normalized inputs $c_{ij}^{norm}$ via learning the covariances $c_{iN}^{norm}$ of rate-normalized inputs with the neuron's own output. We introduce an STDP rule that achieves this by including a component that normalizes the postsynaptic contributions of each presynaptic spike by an increasing function of the recent rate. We call the rule fatiguing STDP (FSTDP), because the rate-normalization component is a synaptic fatigue mechanism, which we combine with STDP.

\begin{figure}[h!]
\hspace*{-0.5cm}
\centering
\begin{tabular}{c}
\includegraphics[width = 3.2in]{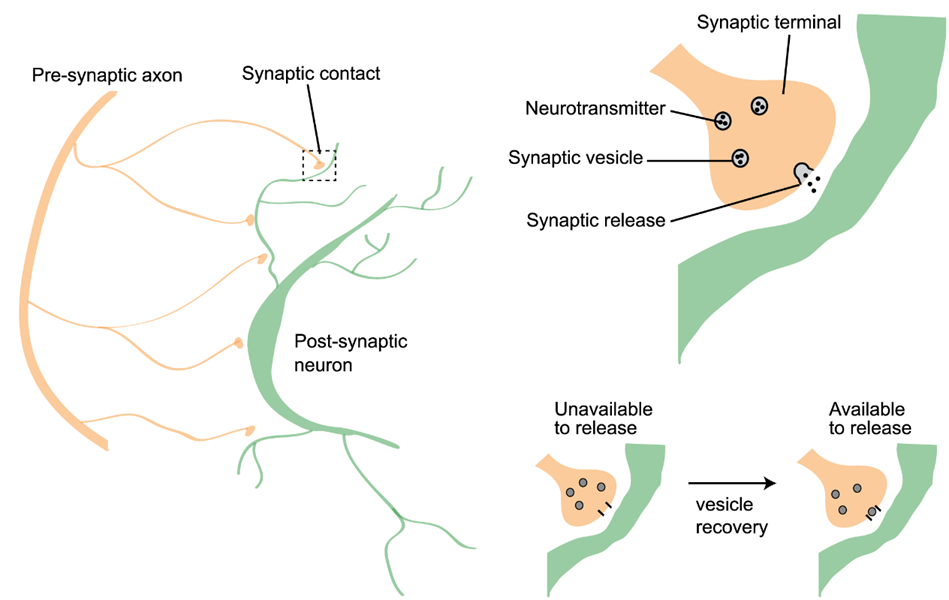}
\end{tabular}
\caption{Synaptic fatigue is strongly motivated by findings that depletion of synaptic neurotransmitter vesicles induces a form of short term
depression in synapses throughout the nervous system (adapted from\cite{Y2012rosenbaumPLOSCB}).} \label{fig:bioinsp}
\end{figure}

Synaptic fatigue is a form of short-term plasticity (STP), specifically short-term depression. It changes the synaptic efficacy in a transient manner. It has been observed in
biological synapses as depletion of the neurotransmitters in the presynaptic terminal by each transmitted spike, and subsequent gradual replenishment (Fig. \ref{fig:bioinsp})
\cite{Tsodyks1997, Y2012rosenbaumPLOSCB}. It has been efficiently implemented in CMOS-based neuromorphic circuits \cite{Qiao_etal15} but not in synapses capable of long-term plasticity.

\begin{figure}[h!]
\hspace*{-0.5cm}
\centering
\begin{tabular}{c}
\includegraphics[width = 3.5in]{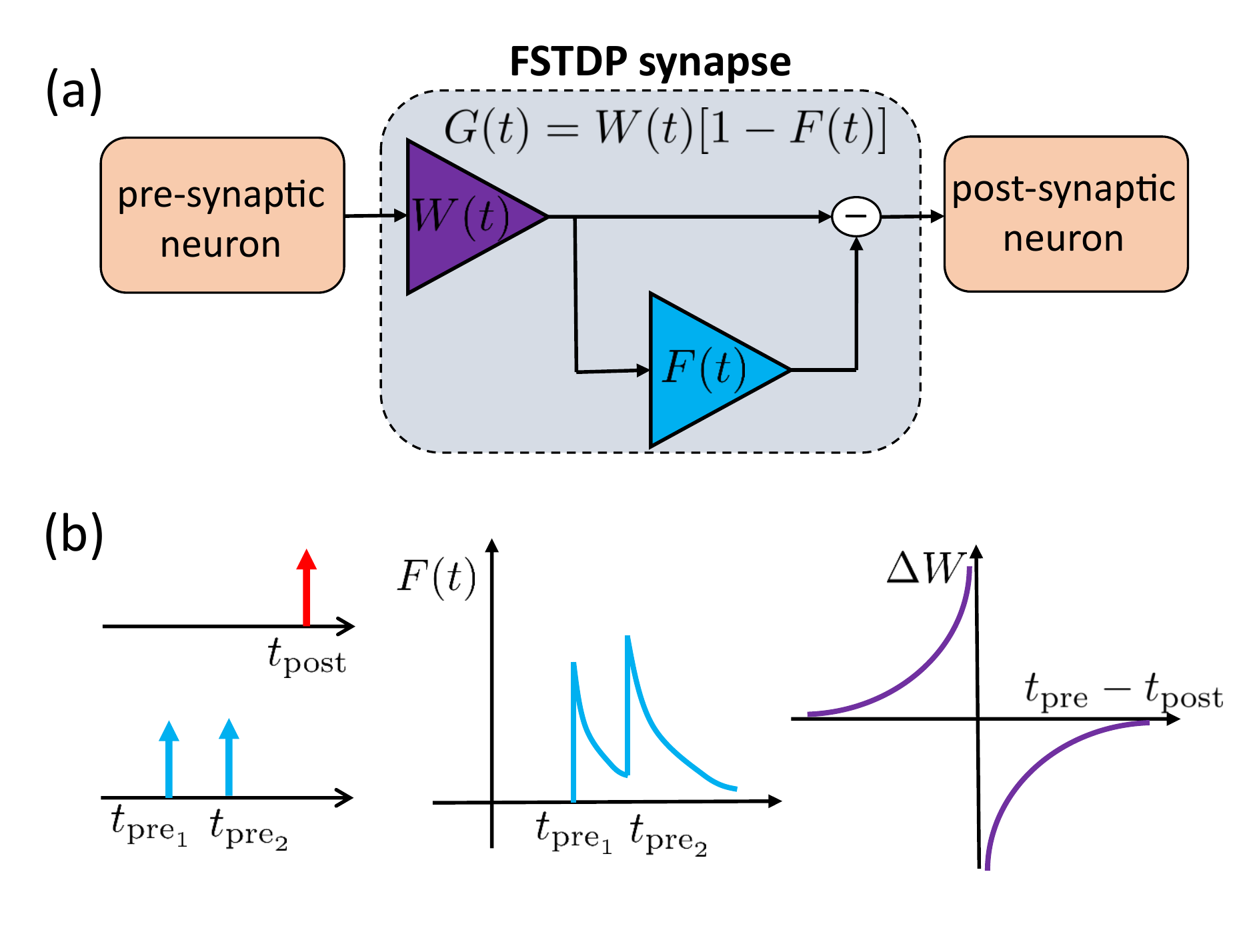}
\end{tabular}
\caption{(a) With the introduction of synaptic fatigue the synaptic efficacy is given by $G(t) = W(t)[1-F(t)]$. (b) Schematic illustration of the
synaptic weight change due to STDP and a characteristic evolution of the $F(t)$ function in response to presynaptic spikes.} \label{fig:fstdp}
\end{figure}

In presence of synaptic fatigue, as shown in Fig. \ref{fig:fstdp}(a), the synaptic efficacy $G(t)$ is given by
\begin{equation}
G(t) = W(t)[1-F(t)]
\end{equation}
where $W(t)$ denotes the stored synaptic weight and $F(t)$ is a function that depends on the time of arrival of the presynaptic spikes. In the
absence of those spikes, $F$ tends to zero as $t\rightarrow \infty$ and thus $G(t) \rightarrow W(t)$. For instance, $F(t)$ can be implemented in a spike-based way as a function that increases its value by a fixed amount upon the arrival of a presynaptic spike and decays exponentially.

Synaptic fatigue increases temporarily when a pre-synaptic input spike arrives, so that if the next spike arrives soon after, i.e. when the instantaneous input rate increases, its postsynaptic effect is decreased. In contrast, STDP is a long-term plasticity mechanism that changes the synaptic weight permanently. The synaptic weight changes depending on the pre- and postsynaptic
spikes, as shown schematically in Fig. \ref{fig:fstdp}(b).


\section{Comparative analytical study of FSTDP and STDP}
In this section, we present an analysis of FSTDP versus STDP. We start from the general conditions that need to be satisfied by the learning rule for a neuron to learn a spike-timing code and show that FSTDP can provide these when STDP cannot. We consider spatio-temporally coded spike trains, which are sequences of spikes that contain temporal correlations between the sequences. In Hebbian learning tasks, a neuron with
$N$ input synapses successfully learns such a code if the weights of the $N_{cor}$ synapses that receive the temporally correlated input sequences are potentiated, and if the $N_{uncor}$ synapses that receive the
uncorrelated sequences are depressed.
This requires that the expected value of the weight updates for temporally correlated synapses be positive, i.e., that the
probability of their potentiation exceed the probability of their depression, and vice versa for uncorrelated synapses:

\begin{numcases}{} E[\Delta W_{cor}]>0 \label{cor}
\\E[\Delta W_{uncor}]<0 \label{uncor}.\end{numcases}

For any synapse $i\in\{1, 2, ..., N\}$ the expected weight update is
\begin{equation}
E[\Delta W_i]=\int_{-\infty}^{\infty}P_i(\Delta t)\cdot k(\Delta t) \, \mathrm{d} \Delta t, \label{EW}
\end{equation}
where $\Delta t = t_\text{pre}-t_\text{post}$ is the difference between presynaptic ($t_\text{pre}$) and postsynaptic ($t_\text{post}$) spike
timings, $P_i(\Delta t)$ is the probability density of this temporal difference, and $k(\Delta t)$ is the temporal kernel of the STDP rule.

$P_{i}(\Delta t)$ is uniform if pre- and postsynaptic spike trains are
uncorrelated: $P_{i0}(\Delta t)=P_0$.
This balance is perturbed by the causal effect of some pre- on postsynaptic spikes.
Specifically, there is an added probability
$\prescript{}{causal}{P}$ that each presynaptic spike of synapse $i$ will bring the postsynaptic neuron to its firing threshold $V_{th}$, causing a
postsynaptic spike. The caused spike occurs after a delay $t_{d}\geq 0$. This delay is given by the temporal dynamics of the excitatory postsynaptic potential (EPSP) caused by the spike. If the EPSP is immediate and instantaneous, there is no delay.
Because of the added $\prescript{}{causal}{P}$,

\begin{equation}
P_i(\Delta t)=m\cdot \left[P_0 + \prescript{}{causal}{P}\cdot\delta(\Delta t + t_{d})\right],
\end{equation}
where $m<1$ is a renormalization factor of the probability density, keeping its integral equal to one after the addition of the causal term; $\delta(x)$ is the Dirac delta function. $m$'s precise value is not essential for the rest of the analysis.
Then, Eq. (\ref{EW}) becomes
\begin{multline}
    E[\Delta W_i]=mP_0\int_{-\infty}^{\infty} k(\Delta t) \mathrm{d} \Delta t+ \prescript{}{causal}{P}\cdot k(-t_{d})\\
    =mP_0\epsilon+\prescript{}{causal}{P}\cdot k(-t_{d}),
\label{potdep}\end{multline} where $\epsilon$ is a factor accounting for any asymmetry between the positive and negative time windows of the STDP kernel,
and is usually chosen to be $\epsilon<0$.

After solving  conditions (\ref{cor}) and (\ref{uncor}), and Eq. (\ref{potdep}) for $\prescript{}{causal}{P}$, the conditions necessary for learning the temporal code become
$\prescript{}{causal}{P}_{uncor}<\displaystyle -\frac{m P_0\epsilon}{k(-t_{d})}<\prescript{}{causal}{P}_{cor}$

\begin{equation}
\Rightarrow\frac{\prescript{}{causal}{P}_{uncor}}{\prescript{}{causal}{P}_{cor}}<1. \label{condition}
\end{equation}
As shown in inequality (\ref{condition}), the ratio between $\prescript{}{causal}{P}_{uncor}$ and $\prescript{}{causal}{P}_{cor}$ is key to
understanding how the learning of the temporal code is facilitated via STDP.

\begin{figure}[h!]
    \centering
    \hspace*{-0.5cm}
    \begin{tabular}{c}
        \includegraphics[width = 3.5in]{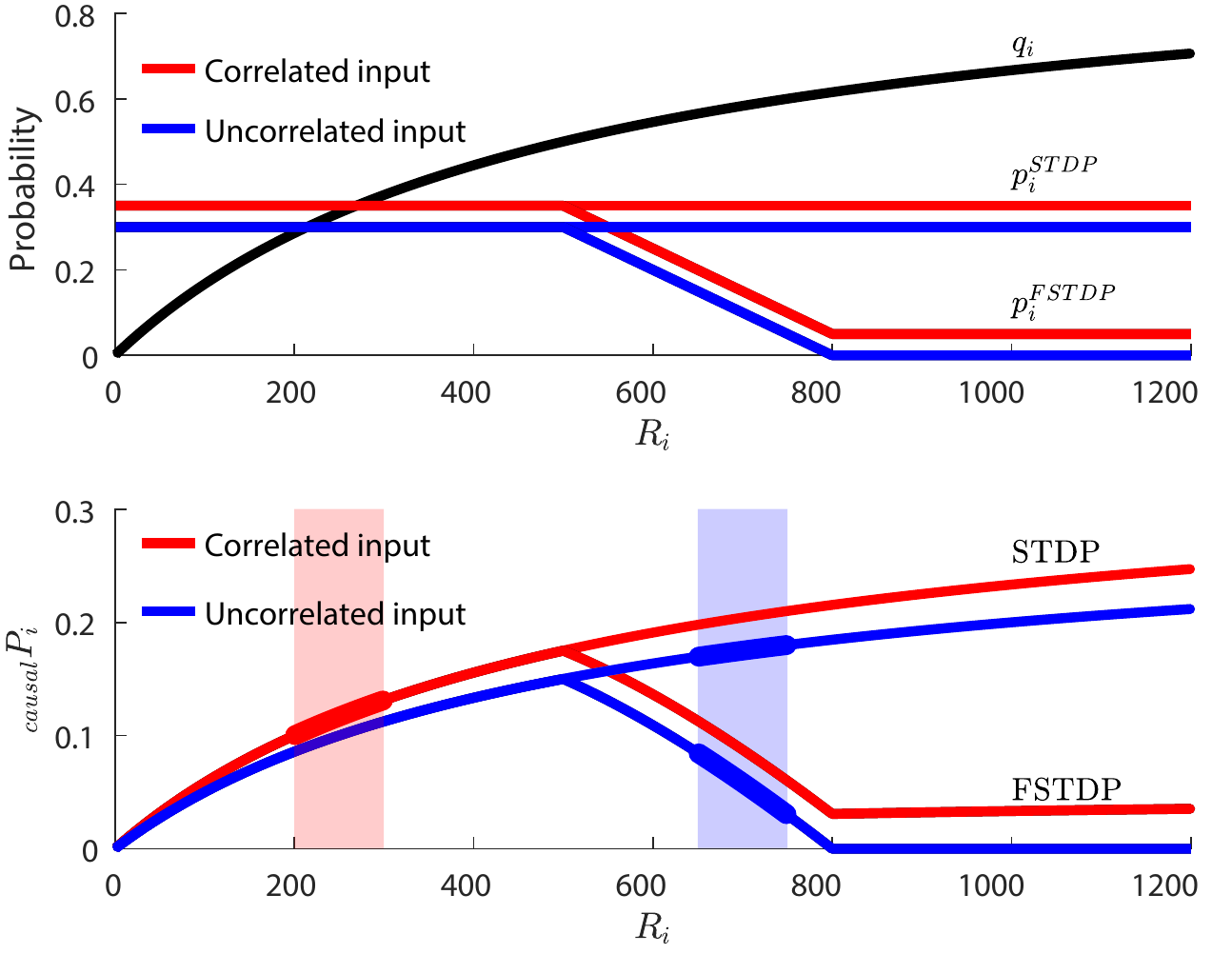}
    \end{tabular}
    \caption{Learning of spike vs rate correlations in FSTDP vs STDP. (a) The probabilities $p_i$ for STDP (flat curves) and FSTDP (branching curves), and $q_i$, as a function of the input rate to a synapse $i$. (b) The probability $\prescript{}{causal}{P}_i=q_i\cdot p_i$ in STDP and FSTDP. In FSTDP, presynaptic spike timing correlations cause postsynaptic spikes, while STDP can be dominated by high rates (e.g. red vs blue shaded regions).} \label{fig:math}
\end{figure}

Let us express $\prescript{}{causal}{P}$ as
$\prescript{}{causal}{P}_i=q_i\cdot p_i$, where $q_i$ is the probability that a presynaptic spike belongs to synapse $i$ from all $N$ synapses. $q_i$ is given by the rate $R_i$ of the input
spikes as a proportion of the sum of all input rates:
\begin{equation}
q_i=\frac{R_i}{\sum\limits_{j=1}^{N}R_j}.
\end{equation}
$q_i$ increases with the rate $R_i$, approaching one (Fig. \ref{fig:math} (a), black line).

$p_i$ is defined as the probability that the presynaptic spike along with its coincidences will bring the neuron to firing threshold given by,
\begin{equation}
p_i=P(V \geq V_{th}-(G_i+n))
\end{equation}
where $V_{th}$ is the postsynaptic neuron's firing threshold, $G_i$
is the efficacy of synapse $i$, and $n$ is a factor accounting for the
excitatory postsynaptic potential caused by spikes from other synapses that coincide with the spike in question because of temporal correlations across the synapses.
$n$ increases with increasing correlation of other synapses
with synapse $i$. Under the assumption of a linear increase of the membrane potential due to the integrated synaptic inputs,
\begin{equation}p_i=\frac{G_i+n}{V_{th}}=p_{i0}+p_{ic},\label{eq:picpi0} \end{equation}
where $p_{i0}=G_i/V_{th}$ and $p_{ic}=n/V_{th}$.

In standard STDP rules, $p_{i0}$ and $p_{ic}$ are independent of $R_i$, as $G_i$ equals the weight $W_i$. Therefore, $p_i^{STDP}(R_i)$ is a constant value (Fig. \ref{fig:math}(a)) and
$\prescript{}{causal}{P}_i =q_i\cdot p_i^{STDP}$ is a rescaled version of $q_i$ which increases with $R_i$ to approach $p_i^{STDP}$. Hence, $\prescript{}{causal}{P}_i$ is higher
for synapses with higher rates, and explains why STDP is sensitive to input firing rates (Fig. \ref{fig:math}(b)).

For rates of correlated inputs that are equal to or higher than uncorrelated ones, $\prescript{}{causal}{P}_i$ is higher for the correlated inputs, in both the STDP and FSTDP cases (Fig. \ref{fig:math}(b), red vs blue lines labeled $STDP$). This is due to the $p_{ic}$ component in Eq. (\ref{eq:picpi0}) which increases as correlations between correlated inputs increase, suggesting that STDP suffices to learn a temporal code in this case. However, when the rates of uncorrelated inputs are much higher than the rates of correlated inputs (Fig. \ref{fig:math} (b), blue- vs red-shaded region respectively), then, in STDP, $\prescript{}{causal}{P}_i$ is higher for the uncorrelated inputs.

In the case of FSTDP, even though $p_{ic}$ is independent of $R_i$, $G_i=W_i\cdot F_i(R_i)$
decreases for $R_i>R_{th}$, where $R_{th}$ is
the rate threshold over which fatigue has an effect and over which $p_{i0}$ also decreases. As a result, $p_i$ decreases for high rates, reaching
its minimum value, $p_{ic}$. Therefore, for FSTDP, $\prescript{}{causal}{P}_i$ is a curve that increases to reach a maximum and then decreases to
reach its minimum, $p_{ic}$.

The rate where $\prescript{}{causal}{P}_i$ is maximum depends on $R_{th}$. $R_{th}$ and the slope of decrease correspond to parameters of the synaptic fatigue component of FSTDP. Specifically in the spike-based implementation of fatigue, the time constant of the fatigue transient determines $R_{th}$, and the increase in fatigue upon each presynaptic spike determines the slope of the decreasing part of the $\prescript{}{causal}{P}_i$ curve. FSTDP can reduce $\prescript{}{causal}{P}_i$ enough for high-rate uncorrelated inputs to satisfy Eq. (\ref{condition}). This increases the sensitivity to spike-timing-correlated inputs relative to spike-timing-uncorrelated inputs even if the latter have high rate correlations.

This analysis is based on the Hebbian forms of STDP, but it applies equally to the anti-Hebbian case by virtue of the symmetry between the two.

\section{Detection of low-rate correlated processes with FSTDP}
\begin{figure*}[h!]
    \centering
    \begin{tabular}{c}
        \includegraphics[width = 4.8in]{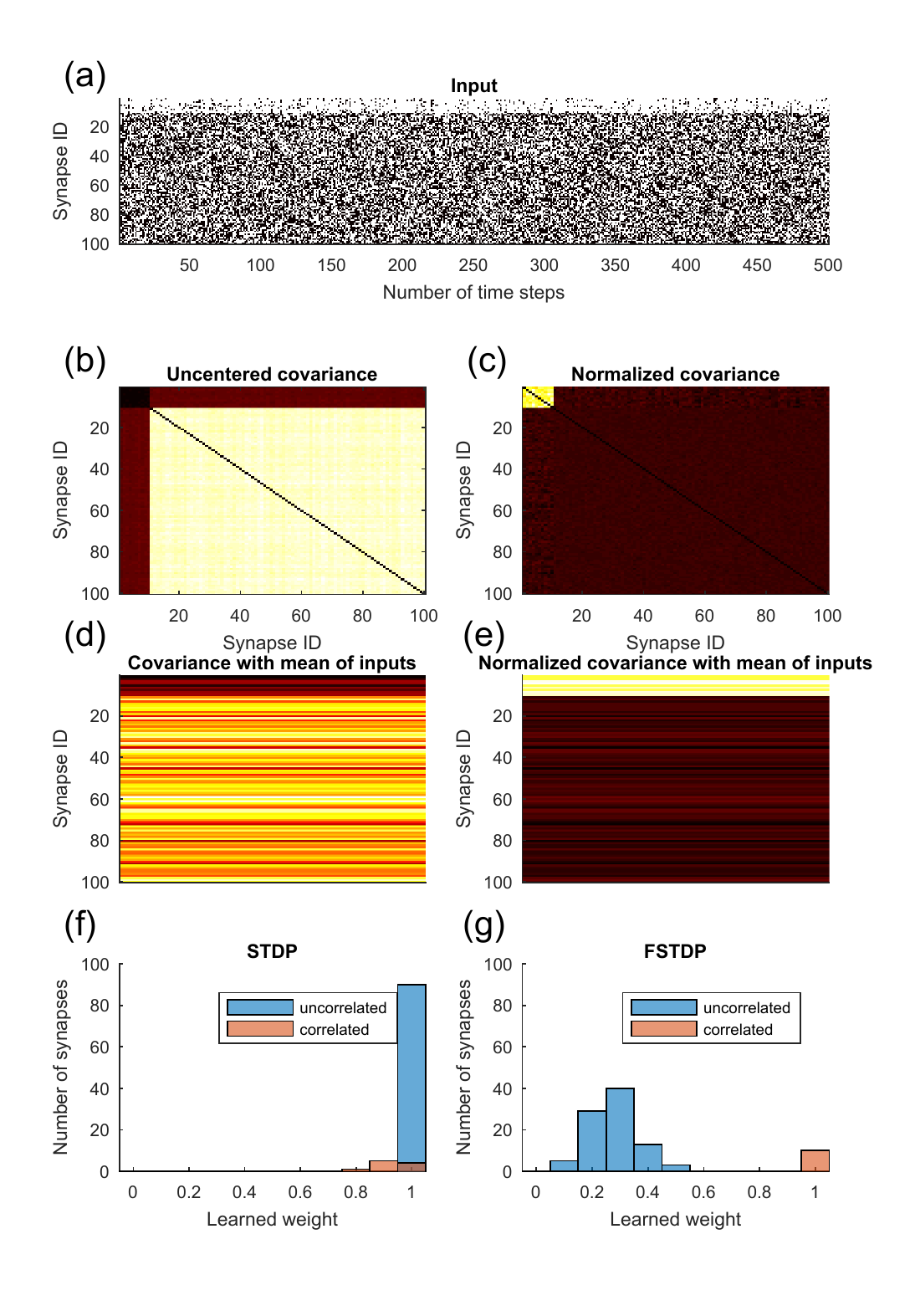}
    \end{tabular}
    \caption{(a) A sample of the synthetic data used. (b) Uncentered covariance matrix. The correlations in the low-rate inputs are obscured by the high rates in the other channels. (c) Normalized
covariance matrix revealing the correlations in the low-rate channels. In both matrices, only the off-diagonal elements are shown. (d) and (e) Covariance and normalized covariance of each input spike train
with the average of all inputs. (f) Weight distribution after learning with STDP. The high uncorrelated rates cause STDP to fail to detect the
correlated inputs. (g) Learning with FSTDP leads to a clear separation of the correlated from the uncorrelated synapses, with the correlated ones being potentiated and the uncorrelated ones being depressed.} \label{fig:sim}
\end{figure*}

In this section, we present a simulation study to investigate the effectiveness of FSTDP in detecting correlated streams of binary events
in real time, under the adverse condition that these correlated streams are characterized with significantly lower frequencies than the rest and than
uncorrelated inputs. In addition, we contrast the learning with FSTDP to learning with STDP. We generated 100 Poisson binary processes, of which 10 are correlated with a correlation coefficient $c=0.1$, and the rest are
uncorrelated. Each of the correlated processes has a mean rate of $R_i=1\mathrm{Hz}, \,i\in[1,10]$, while the uncorrelated processes have a higher mean rate of
$R_j=5\,\mathrm{Hz},\,j\in[11,100]$. A sample of the data is depicted in Fig. \ref{fig:sim}(a); the substantially lower rate of the first 10 processes is
clearly visible.

To validate the results of the STDP and FSTDP algorithms quantitatively, we computed the uncentered ($cov$) and the normalized ($normcov$) covariance matrices of the input streams, respectively (see Section \ref{sec:fstdp}, $c_{ij}$) and used them to define the objective of the correlation detection task that each of the learning rules is best suited for. To predict the outcome of learning, we computed the covariances and the normalized covariances of each input stream with the mean input (see Section \ref{sec:fstdp}, $c_{iN}$).

The network was trained in a completely unsupervised way. In particular, each of the 100 spike trains was provided to one of the 100 synapses of a leaky integrate-and-fire neuron, and the synaptic weights changed according to a pre-defined spike-timing-dependent rule of plasticity. The synaptic weights, which were bounded between zero and one, were initialized to their middle values before training. We trained the network with standard STDP, then we also trained it with FSTDP, and compared the results. The presence or not of synaptic fatigue was the key difference between the two instantiations of the simulation, i.e., the key relevant parameters are, firstly, the jump in synaptic fatigue per presynaptic spike, which was set equal to one, and, secondly, the time constant of the exponential decay of synaptic fatigue, which was set equal to five time steps.

In STDP, high rates drive the output firing and thus also the learning. As a consequence, and as expected from the covariance of the inputs with the mean input (Fig. \ref{fig:sim} (d)) learning through STDP fails to detect the temporally correlated inputs and mostly strengthens the high-rate synapses instead (Fig. \ref{fig:sim} (f)). This result shows that STDP detects the spurious correlations introduced by the high rates, which are salient in the uncentered covariance matrix.

In FSTDP, it is the coincidences across input channels which are rare relative to the corresponding input rates that drive the postsynaptic firing and the learning. As a result, and as predicted by the high values of normalized covariance of the low-rate inputs with the average input (Fig. \ref{fig:sim} (e)), FSTDP leads to successful potentiation of the correlated inputs, matching well both the target set by the normalized covariance matrix (Fig. \ref{fig:sim} (c)) and the coefficients used for generating the data.

These results directly demonstrate the power of FSTDP and its potential to lift limitations of other unsupervised spike-timing-based learning rules.

\section{Demonstration of FSTDP with real-world data sets and neuromorphic hardware}
In this section we present an application of the FSTDP for correlation detection in weather patterns. We also present an implementation of such a
network in a neuromorphic platform where phase-change memory devices serve as synaptic elements.

\begin{figure}[h!]
\centering
\begin{tabular}{c}
\includegraphics[width = 3in]{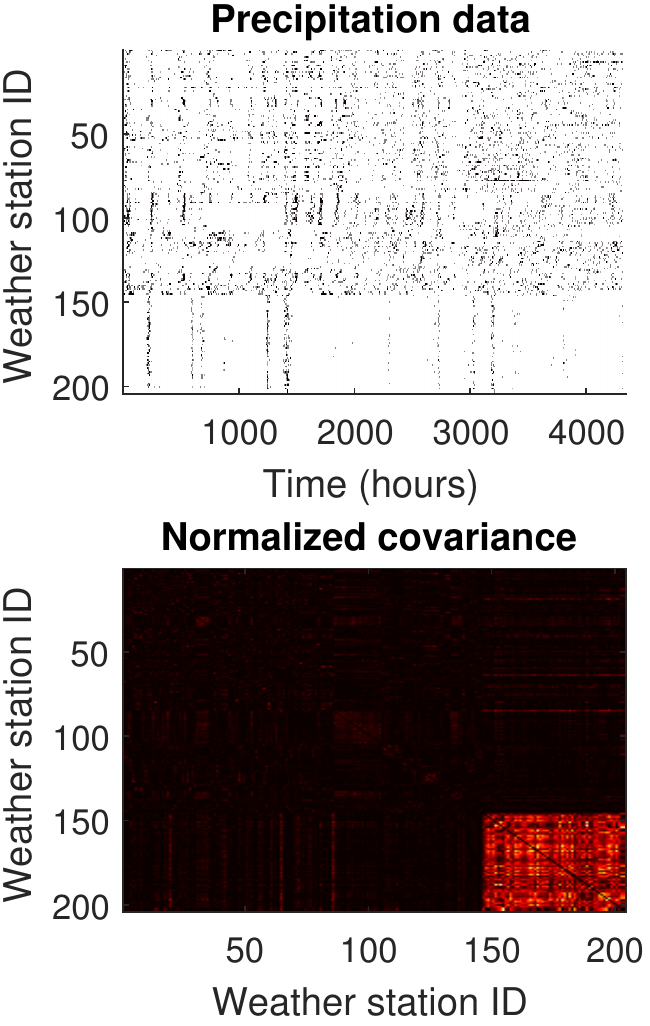}
\end{tabular}
\caption{(a) Depiction of the event-based data streams based on the rainfall information from 205 weather stations across the continental USA. For each station, each hour of rain is marked black. (b) The normalized covariance matrix (off-diagonal elements) visualizes the low-rate but highly correlated weather stations.} \label{fig:weather1}
\end{figure}
The weather data was obtained from the National Oceanic and Atmospheric Administration (http://www.noaa.gov/) database of quality-controlled local
climatological data. It provides hourly summaries of climatological data from approximately 1600 weather stations in the United States of America.
The measurements were obtained over a 6-month period from January 2015 to June 2015 (181 days, 4344 hours). We generated one binary stochastic
process per weather station. If it rained in any given period of 1 hour in a particular geographical location corresponding to a weather station,
then the process takes the value 1; else it will be 0. We selected 205 stations, such that they included 58 locations with scarce but correlated rainfall, and 147 with frequent but uncorrelated rainfall. These groups of stations were found by use of k-means clustering. The event-based data streams are presented in Fig. \ref{fig:weather1}(a). As shown in Fig. \ref{fig:weather1}(b), if we compute the normalized covariance matrix, it is possible to identify these highly correlated weather stations in spite of the low rate of rainfall.
We performed an experiment to obtain the same result using the spiking neural network equipped with FSTDP.

\begin{figure}[h!]
\centering
\begin{tabular}{c}
\includegraphics[width = 2.8in]{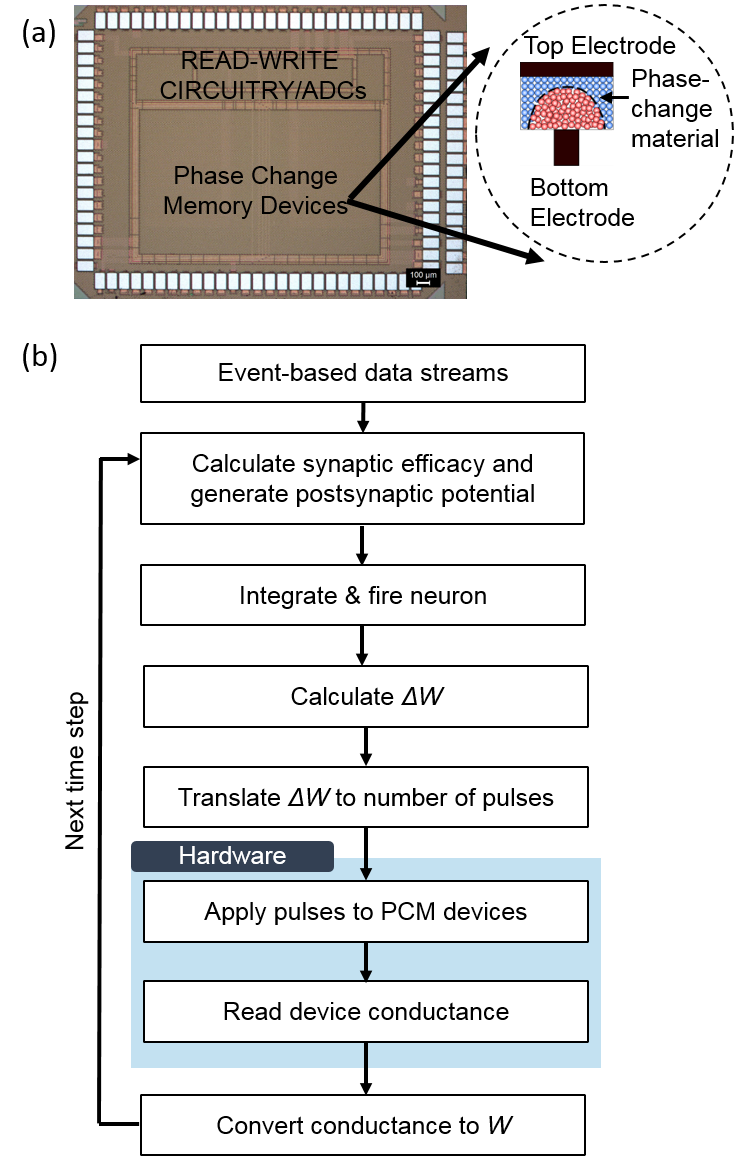}
\end{tabular}
\caption{(a) The phase-change memory (PCM) chip fabricated in 90-nm CMOS technology contains 3 million PCM devices based on doped
Ge$_2$Sb$_2$Te$_5$ (GST) phase-change material. (b) The flow chart of the experiment. Tasks shown in a blue rectangle are done on hardware, whereas the other tasks are done on software.} \label{fig:chip}
\end{figure}

The experiment was performed in a neuromorphic hardware platform based on a phase-change memory (PCM) chip containing 3 million PCM devices. Each
PCM device is accessed by a thin-oxide n-type field-effect transistor (FET). An optical micrograph of the chip and a schematic of the PCM device are
shown in Fig. \ref{fig:chip}(a). The mushroom-type PCM devices are based on doped Ge$_2$Sb$_2$Te$_5$ (GST) and have a sub-lithographically defined
bottom electrode of radius \unit[20]{nm} \cite{Y2007breitwischVLSI}. In addition to the PCM cell array, the chip integrates the circuitry for cell
addressing, an on-chip analog-to-digital-converter (ADC) for cell readout, and voltage- or current-mode cell programming (writing). The chip is
connected to an analog-front-end (AFE) board that contains a number of digital-to-analog-converters (DACs) and ADCs along with discrete electronics,
such as power supplies, voltage and current reference sources. A field-programmable-gate-array (FPGA) board with an embedded processor and Ethernet
connection is used to implement the higher-level routines associated with the experiments. Each synaptic weight is stored in the conductance states
of the PCM devices \cite{Y2016tumaEDL}. The experimental sequence is shown schematically in Fig. \ref{fig:chip}(b).

\begin{figure}[h!]
\centering
\begin{tabular}{c}
\includegraphics[width = 2.8in]{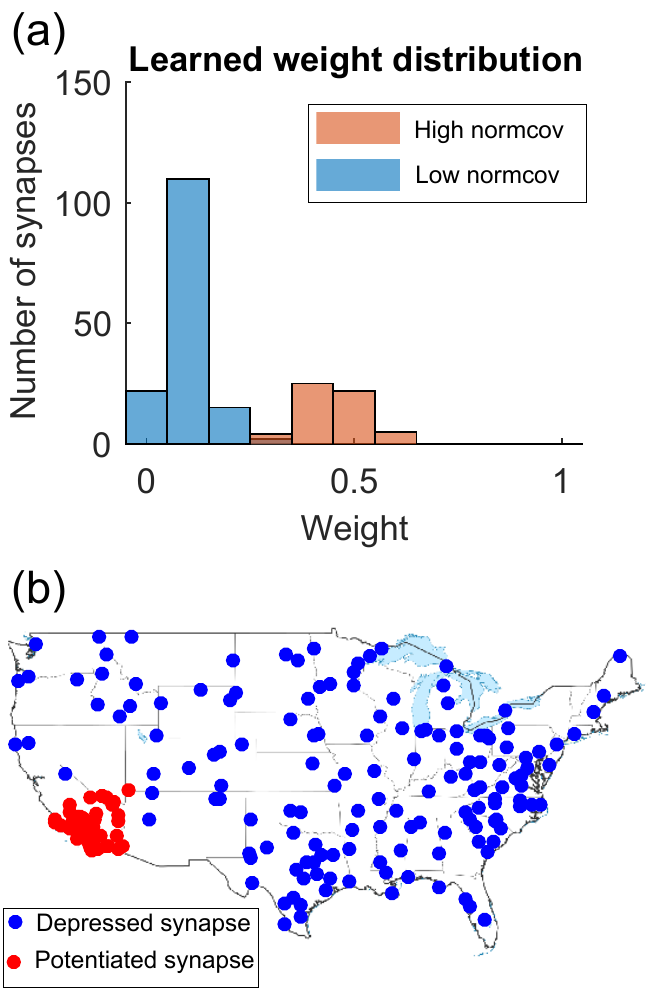}
\end{tabular}
\caption{(a) The histograms of the correlated and uncorrelated synaptic weights (red and blue respectively) after learning show that STDP can detect the correlated weather stations. The synaptic weights are stored in the conductance states of PCM devices in a hardware platform.
(b) Illustration of the geographical location of the weather stations that are found to be temporally correlated (red dots) according to the learned synaptic weights.} \label{fig:weather2}
\end{figure}

The results of the experiments are presented in Fig. \ref{fig:weather2}. The normalized synaptic weights at the end of the experiment are given in
Fig. \ref{fig:weather2}(a). The synaptic weight distribution shows that using FSTDP we are able to identify the weather stations with highly
correlated rainfall. Fig. \ref{fig:weather2}(b) gives the geographical location of the weather stations; the stations with correlated
rainfall are shown in red. As one would expect, the temporal correlations in rainfall also corresponds to geographical proximity. Moreover, these
weather stations are located in California, which is known to have relatively low levels of precipitation compared to other regions of the US. These
experiments highlight the applicability of this type of learning algorithms on realistic data sets as well as the feasibility of neuromorphic
hardware with memristive synapses to realize the learning in-situ in an efficient manner. Even though in this particular demonstration, we used the
PCM devices mostly for synaptic weight storage and weight updates, one could foresee the development of memristive devices in which also the short-term
synaptic dynamics could be realized using the device physics \cite{Y2016wangNatMat}.

\section{Discussion}
We presented, analyzed, and demonstrated the effectiveness of FSTDP, i.e., a learning rule for spiking neural networks that allows learning of temporal codes such as spike-timing codes in the presence of slower codes such as rate codes. In this regard, FSTDP compares favorably to other unsupervised spike-based learning rules, which face problems in learning the fine spatiotemporal structure in the spiking inputs, as we show theoretically and experimentally. The rule is local and event-based.

Despite the spatially and temporally local event-based nature of the rule, it gives rise to global and long-term learning effects to a network that uses it. This lends it an efficiency that is well-aligned with the scope of SNNs. The LTP dynamics emerge from the combination of an STP with an LTP event-based component. Despite this simplicity, we show that it overcomes the significant limitations that other unsupervised learning rules for SNNs face in learning from spike-timing codes in the presence of rate codes. Other potential approaches would only address the issue partly. For instance, setting the postsynaptic neuron's time constant of leak to match the time constant of the temporal correlations, would not be able to address high enough firing rates that introduce uncorrelated spikes in comparable intervals. Alternatively, probabilistically down-sampling the spikes in high-rate input channels would reduce the postsynaptic effect of high-rate channels, but would also collaterally omit some of the spikes that form the spike-timing code, hindering the task of learning from them. Another alternative mechanism that would target the same learning issue would be the introduction of fatigue, not on the synaptic efficacy as in FSTDP, but directly on the weight updates, such that weight updates are smaller if they are frequent. However, this would generically prevent learning in high-rate channels, disallowing their potentiation and depression regardless of any spike-timing correlations with other channels.

One of the potential advantages of SNNs over traditional ANNs is the use of spikes, i.e., binary events. In this type of neural networks, time encodes itself in the timing of the events, so they can learn and operate in real time. Spike timings can embed information in the neural code additional to the mean firing rate, effectively increasing the bandwidth of communication between neurons in the network compared with a single analog value which as in traditional ANNs. Furthermore, because of their asynchronous and event-based nature, combined with the dual role of synapses as both memory and computational elements, SNNs are well-suited for efficient hardware implementations of massively-parallel post-von-Neumann computing architectures. However, their potential advantages that stem from the use of spiking events are hard to exploit in real-world applications, largely due to the lack of algorithms that can effectively learn to extract the useful information from the arbitrary complexity of event-based codes. This is particularly true when the structure underlying the complexity of the training data is unknown to the network's user, so that the algorithm must learn in an unsupervised manner. FSTDP addresses a significant source of  such complexity in the neural code, which results from the differences in the timescales between useful and distracting information embedded in spatiotemporal spike patterns. This paves the way to applications of unsupervised learning from real-world temporally-coded data, which are rarely isolated from rate codes.

Furthermore, it may well be that variations of FSTDP are employed by the biological nervous system. Firstly, both spike-timing-dependent plasticity and synaptic fatigue have been experimentally observed in synapses connecting biological neurons. Moreover, the brain has likely evolved to exploit the large potential information content in fast-varying temporal codes. These codes need to operate in the presence of slower temporal codes such as varying firing rates encoding sensory stimuli or motor outputs, large-scale neural activity rhythms, and state-dependent responses to stimuli. Based on the existence of FSTDP's constituent elements and the brain's likely need to learn from temporal codes in the presence of rate codes, it is plausible that FSTDP is used by the brain.

A limitation of the work we present here concerns our implementation of FSTDP with neuromorphic hardware. The storage of the weights, the weighted transmission of presynaptic spikes to the postsynaptic neuron, and the updates of the stored weight are all accommodated by the neuromorphic hardware. However, the overall synaptic efficacy is a hybrid of software and hardware, as the fatigue component of the synaptic dynamics is simulated in software. The effectiveness and computational power of FSTDP may motivate the development of future pure hardware FSTDP implementations.

A second limitation is that our experimental demonstrations do not cover the full scope of FSTDP. For instance, FSTDP could be used not only for correlation detection, but also in unsupervised classification, sequence generation, dimensionality reduction, and other tasks. Furthermore, the network we used comprised a single postsynaptic neuron, but certain applications may require more complex network architectures. Nevertheless, the successful detection of temporal correlations at the synaptic level is the computational primitive of all these tasks and relevant network architectures, and was clearly demonstrated in the two examples. The rule can handle any type of data that is encoded as spike trains that include spatiotemporal correlations.

In conclusion, our analyses and demonstrations of the FSTDP learning rule suggest that it may enable the efficient use of software and hardware SNNs in a variety of applications of learning from real-world temporal data.

\section*{Acknowledgments}
We would like to thank the PCM team at IBM Research - Zurich and in particular Nikolaos Papandreou for technical assistance, and Lorenz M\"{u}ller
from the Institute of Neuroninformatics (UZH/ETHZ) for an insightful discussion. We would also like to acknowledge partial financial
support from the Swiss National Science Foundation.

\bibliographystyle{IEEEtran}
\bibliography{References}

\begin{thebibliography}{10}
\providecommand{\url}[1]{#1}
\csname url@samestyle\endcsname
\providecommand{\newblock}{\relax}
\providecommand{\bibinfo}[2]{#2}
\providecommand{\BIBentrySTDinterwordspacing}{\spaceskip=0pt\relax}
\providecommand{\BIBentryALTinterwordstretchfactor}{4}
\providecommand{\BIBentryALTinterwordspacing}{\spaceskip=\fontdimen2\font plus
\BIBentryALTinterwordstretchfactor\fontdimen3\font minus
  \fontdimen4\font\relax}
\providecommand{\BIBforeignlanguage}[2]{{%
\expandafter\ifx\csname l@#1\endcsname\relax
\typeout{** WARNING: IEEEtran.bst: No hyphenation pattern has been}%
\typeout{** loaded for the language `#1'. Using the pattern for}%
\typeout{** the default language instead.}%
\else
\language=\csname l@#1\endcsname
\fi
#2}}
\providecommand{\BIBdecl}{\relax}
\BIBdecl

\bibitem{Y2015lecunNature}
Y.~LeCun, Y.~Bengio, and G.~Hinton, ``Deep learning,'' \emph{Nature}, vol. 521,
  no. 7553, pp. 436--444, 2015.

\bibitem{Y1997maassNN}
W.~Maass, ``Networks of spiking neurons: the third generation of neural network
  models,'' \emph{Neural Networks}, vol.~10, no.~9, pp. 1659--1671, 1997.

\bibitem{Y2014merollaScience}
P.~A. Merolla \emph{et~al.}, ``A million spiking-neuron integrated circuit with
  a scalable communication network and interface,'' \emph{Science}, vol. 345,
  pp. 668--673, 2014.

\bibitem{Y2015indiveriIEEEProc}
G.~Indiveri and S.-C. Liu, ``Memory and information processing in neuromorphic
  systems,'' \emph{Proceedings of the IEEE}, vol. 103, no.~8, pp. 1379--1397,
  2015.

\bibitem{Y2016tumaNatNano}
T.~Tuma, A.~Pantazi, M.~Le~Gallo, A.~Sebastian, and E.~Eleftheriou,
  ``Stochastic phase-change neurons,'' \emph{Nature Nanotechnology}, vol.~11,
  pp. 693--699, 2016.

\bibitem{Y2003gutigJNeuro}
R.~G\"{u}tig, R.~Aharonov, S.~Rotter, and H.~Sompolinsky, ``Learning input
  correlations through nonlinear temporally asymmetric hebbian plasticity,''
  \emph{Journal of Neuroscience}, vol.~23, pp. 3697--3714, 2003.

\bibitem{Kempter1999}
R.~Kempter, W.~Gerstner, and J.~{Van Hemmen}, ``{Hebbian learning and spiking
  neurons},'' \emph{Physical Review E}, vol.~59, no.~4, pp. 4498--4514, 1999.

\bibitem{Gilson2011}
M.~Gilson, T.~Masquelier, and E.~Hugues, ``{STDP allows fast rate-modulated
  coding with Poisson-like spike trains},'' \emph{PLoS Computational Biology},
  vol.~7, no.~10, 2011.

\bibitem{Tsodyks1997}
M.~V. Tsodyks and H.~Markram, ``{The neural code between neocortical pyramidal
  neurons depends on neurotransmitter release probability},'' \emph{Proceedings
  of the National Academy of Sciences}, vol.~94, no.~2, pp. 719--723, 1997.

\bibitem{Y2012rosenbaumPLOSCB}
R.~Rosenbaum, J.~Rubin, and B.~Doiron, ``Short term synaptic depression imposes
  a frequency dependent filter on synaptic information transfer,'' \emph{PLoS
  Computational Biology}, vol.~8, no.~6, p. e1002557, 2012.

\bibitem{Qiao_etal15}
N.~Qiao, H.~Mostafa, F.~Corradi, M.~Osswald, F.~Stefanini, D.~Sumislawska, and
  G.~Indiveri, ``{A reconfigurable on-line learning spiking neuromorphic
  processor comprising 256 neurons and 128K synapses},'' \emph{Frontiers in
  Neuroscience}, vol.~9, no. 141, pp. 1--17, 2015.

\bibitem{Y2007breitwischVLSI}
M.~Breitwisch, T.~Nirschl, C.~Chen, Y.~Zhu, M.~Lee, M.~Lamorey, G.~Burr,
  E.~Joseph, A.~Schrott, J.~Philipp \emph{et~al.}, ``Novel
  lithography-independent pore phase change memory,'' in \emph{IEEE Symposium
  on VLSI Technology}.\hskip 1em plus 0.5em minus 0.4em\relax IEEE, 2007, pp.
  100--101.

\bibitem{Y2016tumaEDL}
T.~Tuma, M.~Le~Gallo, A.~Sebastian, and E.~Eleftheriou, ``Detecting
  correlations using phase-change neurons and synapses,'' \emph{IEEE Electron
  Device Letters}, vol.~37, no.~9, pp. 1238--1241, 2016.

\bibitem{Y2016wangNatMat}
Z.~Wang, S.~Joshi, S.~E. Savel/'ev, H.~Jiang, R.~Midya, P.~Lin, M.~Hu, N.~Ge,
  J.~P. Strachan, Z.~Li, Q.~Wu, M.~Barnell, G.-L. Li, H.~L. Xin, R.~S.
  Williams, Q.~Xia, and J.~J. Yang, ``{Memristors with diffusive dynamics as
  synaptic emulators for neuromorphic computing},'' \emph{Nature Materials},
  vol. advance online publication, sep 2016.

\end{thebibliography}

\end{document}